\documentclass{article}

\usepackage{microtype}
\usepackage{graphicx}
\usepackage{booktabs} 
\usepackage{makecell}
\usepackage{authblk}
\usepackage{afterpage}
\usepackage{color, colortbl}
\usepackage{multirow}
\usepackage{array}
\usepackage{xcolor}
\usepackage{rotating}
\usepackage{amssymb}
\usepackage{pifont}
\usepackage{xspace}
\usepackage{subcaption}
\newcommand{\xmark}{\ding{55}}%
\usepackage{nicefrac}       
\usepackage{amsmath}

\usepackage{hyperref}

\usepackage[accepted]{icml2024}

\usepackage{amsmath}
\usepackage{amssymb}
\usepackage{mathtools}
\usepackage{amsthm}

\usepackage[capitalize,noabbrev]{cleveref}

\newcommand{\grad}{\texttt{grad-CROP}}
\newcommand{\att}{\texttt{att-CROP}}

\newcommand{\clip}{\texttt{clip-CROP}}
\newcommand{\sam}{\texttt{sam-CROP}}
\newcommand{\yolo}{\texttt{yolo-CROP}}
\newcommand{\hc}{\texttt{human-CROP}}

\makeatletter
\DeclareRobustCommand\onedot{\futurelet\@let@token\@onedot}
\def\@onedot{\ifx\@let@token.\else.\null\fi\xspace}

\def\eg{\emph{e.g}\onedot} 
\def\ie{\emph{i.e}\onedot}

\makeatletter

\theoremstyle{plain}

\theoremstyle{definition}

\theoremstyle{remark}

\usepackage[textsize=tiny]{todonotes}

\icmltitlerunning{Towards Perceiving Small Visual Details in Zero-shot Visual Question Answering with Multimodal LLMs}

\begin{document}

\twocolumn[
\icmltitle{Towards Perceiving Small Visual Details \\
in Zero-shot Visual Question Answering with Multimodal LLMs}




\begin{icmlauthorlist}
\icmlauthor{Jiarui Zhang}{usc}
\icmlauthor{Mahyar Khayatkhoei}{usc}
\icmlauthor{Prateek Chhikara}{usc}
\icmlauthor{Filip Ilievski}{uv}
\end{icmlauthorlist}

\icmlaffiliation{usc}{University of Southern California, Los Angeles, California, USA}
\icmlaffiliation{uv}{Vrije Universiteit Amsterdam, Amsterdam, Netherlands}

\icmlcorrespondingauthor{Jiarui Zhang}{jzhang37@usc.edu}

\icmlkeywords{Machine Learning, ICML}
\vskip 0.3in
]



\printAffiliationsAndNotice{}  

\begin{abstract}
Multimodal Large Language Models (MLLMs) have recently achieved promising zero-shot accuracy on visual question answering (VQA)---a fundamental task affecting various downstream applications and domains. Given the great potential for the broad use of these models, it is important to investigate their limitations in dealing with different image and question properties. In this work, we investigate whether MLLMs can perceive small details as well as large details in images. In particular, we show that their zero-shot accuracy in answering visual questions is very sensitive to the size of the visual subject of the question, declining up to 46\% with size. Furthermore, we show that this effect is causal by observing that human visual cropping can significantly mitigate their sensitivity to size. Inspired by the usefulness of human cropping, we then propose five automatic visual cropping methods---leveraging either external localization models or the decision process of the given MLLM itself---as inference time mechanisms to improve the zero-shot performance of MLLMs. We study their effectiveness on four popular VQA datasets, and a subset of the VQAv2 dataset tailored towards fine visual details. Our findings suggest that MLLMs should be used with caution in detail-sensitive VQA applications, and that visual cropping is a promising direction to improve their zero-shot performance. To facilitate further investigation of MLLMs' behaviors, our code and data are publicly released \href{https://github.com/saccharomycetes/vicrop}{here}.
\end{abstract}

\begin{figure*}[t]
    \centering
    \includegraphics[trim=0 0 0 0, clip, width=0.99\textwidth]{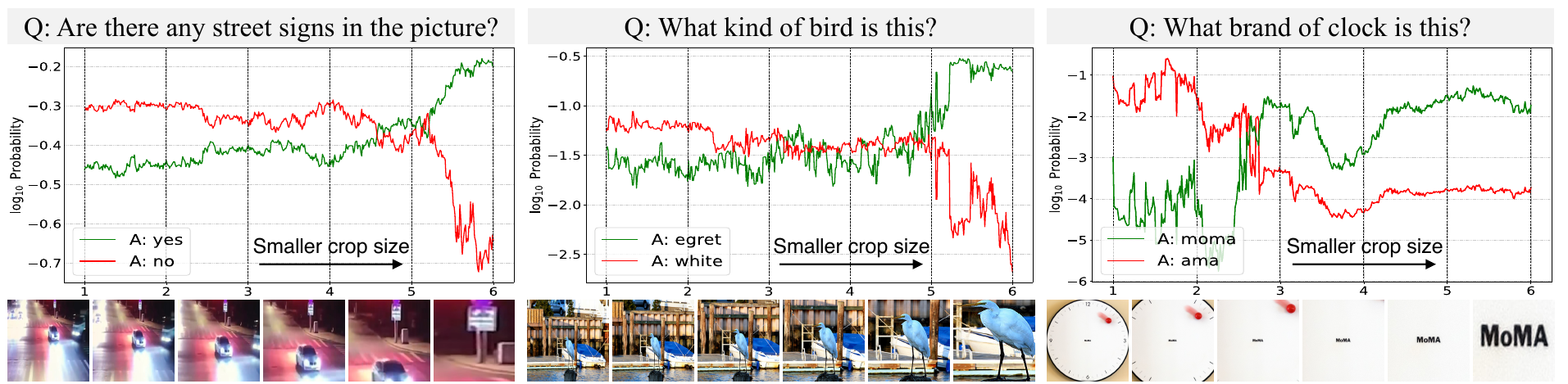}
    \caption{The effect of visual cropping on the probability of answers predicted by BLIP-2 FlanT5$_\mathrm{XL}$ zero-shot VQA model. The x-axis represents the relative crop size around the relevant visual subject of the question (x-axis labels are indices to the respective cropped images displayed under each plot that the model sees at each step). The model gradually finds the correct answer as visual cropping allows it to look closer and thereby better perceive small visual details.}
    \label{fig:crop_fig}
\end{figure*}
\section{Introduction}
\label{sec:intro}

Visual Question Answering (VQA) is a fundamental task with a broad range of downstream applications in many critical domains, from biomedicine~\citep{seenivasan2022surgicalvqa, naseem2022pathvqa} to traffic monitoring~\citep{xu2021sutd, zhang2023study} and remote sensing~\cite{sarkar2021vqaaid, lobry2020rsvqa}. Zero-shot VQA---answering visual questions in a domain without having access to annotated data from that task and domain---is of particular interest since collecting reliable answers for an extensive number of question-image pairs is expensive and time-consuming, and thus impractical for many downstream tasks due to the lack of access to experts, as well as privacy and security concerns~\cite{zhang2023practicevqa}. Recently, Multimodal Large Language Models (MLLMs)~\cite{li2023blip, openai2023gpt4} have shown promising accuracy in zero-shot VQA, commonly attributed to their pretraining on terabytes of image and language data with billion-parameter Transformer-based neural networks. Given their potentially broad adoption in downstream tasks, it is crucial to study their limitations in dealing with various phenomena in images and questions, which is understudied in previous research. To that end, in this work, we investigate whether their question-answering ability is affected by the size of the visual object of interest.

In~\cref{fig:crop_fig}, we provide three motivating examples to illustrate a limitation in MLLMs that we will study in this paper in more detail. 
In these examples, we ask BLIP-2 FlanT5$_\mathrm{XL}$~\cite{li2023blip}, a state-of-the-art MLLM in zero-shot VQA, three questions about relatively small objects in the image,~\ie, questions concerning \textit{small visual details}. 
In the absence of any prior empirical evidence, one might reasonably believe that the accuracy is not significantly affected by the size of the question's visual subject because of the large representational capacity of MLLMs and their pretraining on a large variety of images containing objects of various sizes. 
Contrary to this belief, in~\cref{fig:crop_fig} (left), we observe that initially the model does not recognize the existence of a small street sign and assigns a lower probability to the correct answer; however, zooming into the image towards the street sign gradually increases the probability assigned to the correct answer, suggesting that the model gradually perceives more and more relevant details of the street sign. Similarly, in~\cref{fig:crop_fig} (middle), we observe further evidence of this limitation in perceiving visual details. The model initially predicts \textit{white} as the type of the bird; however, when we zoom into the image towards the bird via visual cropping, without changing the question in any way, we observe that the model gradually assigns higher probability to the correct bird type of \textit{egret}, suggesting that the model was not making a semantic error of misunderstanding what \textit{type} means, rather it was unable to perceive sufficient details to discriminate egret from other white birds, which is mitigated by visual cropping. Similarly, in~\cref{fig:crop_fig} (right), we observe that the model’s initial answer is not entirely irrelevant (ama), suggesting that the model knows where to look based on the question but cannot perceive small visual details, which is again mitigated by visual cropping. This observation is particularly surprising since the visual encoding in BLIP-2 is theoretically not restricted in its visual resolution and therefore should be able to perceive the traffic sign, recognize the bird type, and read text regardless of their relative visual sizes.

The main goal of this paper is to investigate the extent of the limitation observed in~\cref{fig:crop_fig}, and explore potential solutions to mitigate its consequences. In~\cref{sec:human_crop}, we will quantitatively show that there indeed exists a bias against small visual details in MLLMs. Our findings are consistent with concurrent work on evaluating the text-image matching in vision-language joint embedding models, which have observed a reverse correlation between visual object size in images and the text-image matching score~\cite{vlchecklist}, but we further provide an intervention study---manipulating images directly through cropping---to illustrate the causal relationship between object size and MLLM's ability to perceive objects in question answering. In~\cref{sec:vicrop}, we will construct five automatic cropping methods---leveraging either external localization models or the decision process of the given MLLM itself---as potential inference time solutions to the observed bias. Due to computational constraints, our study and proposed methods will be primarily focused on the open-source BLIP-2 MLLM as a proof-of-concept for the utility of visual cropping. Nonetheless, we expect the findings and methods to carry over to other MLLMs to the extent that their visual backbone is similar to BLIP-2, but we leave an extensive application of visual cropping to other MLLMs to future work. To facilitate such future research, we will make the code and models for all methods and experiments publicly available upon publication.

\section{Related Works}
\label{sec:related_works}

\textbf{Multimodal Large Language Models~(MLLMs).}
MLLMs are designed as foundation models that can perform various downstream language and image tasks. These models can be broadly grouped into two categories: \emph{end-to-end pretrained} models, and \emph{modular pretrained} models. The former group includes architectures that are explicitly designed for processing joint image and language data, most notably, the dual-encoder~\cite{clip}, the fusion-encoder~\cite{li2021align-before-fuse}, the encoder-decoder~\cite{cho2021unifying}, and the unified transformer~\cite{wang2022image-as-foreign-lang}, which are trained with common pretraining objectives: image-text matching, and contrastive and masked language modeling. The latter group aims to overcome the expensive pretraining cost of the former group by learning to adapt existing pretrained models: some models use a frozen image encoder and finetune a large language model (LLM) with the pretraining objectives~\cite{zhai2022lit, zhang2021vinvl}, whereas others freeze the LLM instead and finetune the vision encoder with additional adaptor layers~\cite{alayrac2022flamingo, tsimpoukelli2021frozen}. Notably, BLIP-2~\cite{li2023blip} freezes both the vision encoder and the LLM, and directly learns a transformer-based module (denoted Q-Former) on pretraining objectives to bridge the modality gap of its frozen underlying models. Our work will contribute to a better understanding of the sensitivity of MLLMs to image properties, improving their safe and effective use in practice.

\textbf{Visual Localization Methods.}
Dedicated visual localization techniques, such as YOLO~\cite{yolo}, SAM~\cite{kirillov2023segment}, GLIP~\cite{glip}, rely heavily on rich spatial annotations to identify salient regions in images. In contrast, native visual localization techniques, such as grad-cam~\cite{gradcam}, try to localize the salient image region by tracking the gradients of the convolutional classifier's own decision, without requiring any need for spatial annotation. Recent works, PNP-VQA~\cite{pnpvqa} and Img2LLM~\cite{pnpvqa2}, have successfully applied grad-cam to the Transformer structure, identifying the most relevant image patches from BLIP~\cite{blip} model's vision transformer (ViT) by tracking the image-text similarity. Recently, the V* algorithm proposed by~\citep{v-star} enables visual research to enhance MLLM's performance on questions requiring visual details. In addition, visual-based programming techniques~\cite{suris2023vipergpt, visualprogramming} inherit the code capability of LLMs and use them as controllers to call different visual localization models, such as object detection. In this work, rather than proposing a new MLLM or a method, we will provide evidence of the struggle of MLLM to answer questions about small visual details, and further show that this difficulty can be mitigated by employing both dedicated (external) and native visual localization techniques as inference time mechanisms.
\begin{table}[t]
\caption{Sensitivity of zero-shot accuracy of VQA models to the size of visual concepts in TextVQA. As the relative visual size of the answer decreases (right to left in each row), we observe a significant decline in the accuracy of the original models, whereas visual cropping reduces this accuracy gap.\\}
\label{tab:bbox_size}
\centering
\setlength{\tabcolsep}{2pt} 
\begin{tabular}{lcccc}
\toprule
\multirow{2}{*}{Model} & \multirow{2}{*}{Crop Method} & \multicolumn{3}{c}{Answer Bbox Size ($S$)}\\
\cmidrule(lr){3-5}
& & \texttt{small} & \texttt{medium} & \texttt{large} \\
\midrule
BLIP-2  & w/o cropping & 19.91 & 29.07 & 36.81 \\
 FlanT5$_\mathrm{XL}$& \hc & 32.06 & 41.31 & 38.84 \\ \midrule
 
BLIP-2 & w/o cropping & 19.38 & 26.09 & 33.28 \\
           OPT$_\mathrm{2.3B}$            & \hc          & 27.19 & 34.36 & 33.25 \\
\bottomrule
\end{tabular}

\end{table}
\begin{figure*}[t]
    \centering
    \includegraphics[trim=0 0 0 0, clip, width=0.99\textwidth]{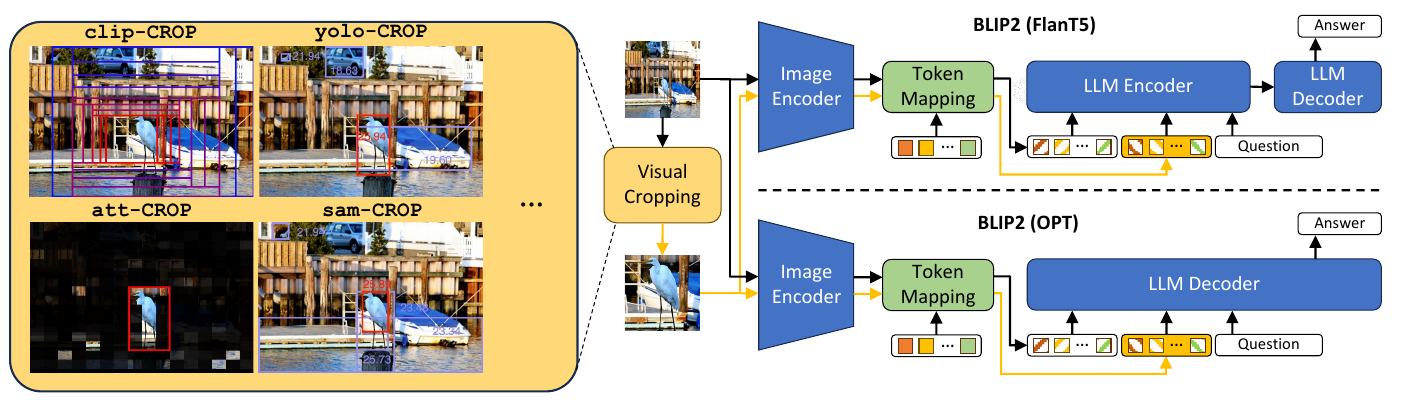}
    \caption{Illustration of the proposed visual cropping approach applied to two variants of BLIP2.}
    \label{fig:methods}
\end{figure*}

\section{Sensitivity of Zero-Shot VQA Models to the Size of Visual Concepts}
\label{sec:human_crop}

In this section, our goal is to quantitatively test our qualitative observations in~\cref{fig:crop_fig}, \ie, the zero-shot VQA models struggle with describing small visual details in images. To that end, we consider the TextVQA dataset, in which for each question we can find the ground-truth bounding box containing the correct textual answer (detailed in~\cref{sec:experiments}). We partition its validation set into three groups based on the relative size of the ground truth bounding box $S = \frac{A_{bb}}{A_{total}}$, where $A_{bb}$ denotes the area of the answer bounding box, and $A_{total}$ denotes the total area of the image: 1) $S<0.005$ (\texttt{small}) consisting of 2822 question-image pairs, 2) $0.005\leq S<0.05$ (\texttt{medium}) consisting of 1833 question-image pairs, and 3) $S\geq 0.05$ (\texttt{large}) consisting of 345 question-image pairs. If a model's perception is not sensitive to the size of visual concepts, we expect it to have similar accuracy in all three partitions. In~\cref{tab:bbox_size}, we observe that the accuracy of both BLIP-2 variants declines across the three groups as the answer bounding box becomes smaller (see \texttt{w/o cropping} rows). BLIP-2 FlanT5$_\mathrm{XL}$ exhibits an accuracy decline of $45.91\%$ from the largest visual concepts group to the smallest visual concepts group, and BLIP-2 OPT$_\mathrm{2.3B}$ exhibits a similar decline of $41.77\%$. These findings show that both models answer questions about visual concepts more accurately when their relative size is larger, \ie, they struggle to perceive fine visual details.

Furthermore, to confirm that the issue is causally related to the size of the visual concept, we conduct an intervention study, where we provide the models with visually cropped images based on the ground-truth bounding boxes, denoted as \hc{}. We observe in~\cref{tab:bbox_size} that \hc{} significantly improves the accuracy of all three models, but more importantly, under \hc{} the accuracy across the three groups becomes more similar: the decline between the \texttt{large} group and the \texttt{small} group is significantly less than that of without cropping setting for all models. This suggests that the perception limitation is indeed caused by the size of the visual concepts, and that visual cropping can mitigate this limitation.

\section{Visual Cropping}
\label{sec:vicrop}

The accuracy gain achieved by human visual cropping in~\cref{sec:human_crop} shows the potential benefit of zooming in towards the relevant region of the input image for improving the accuracy of the zero-shot VQA models. To realize this potential in practice, in this section, we develop automatic visual cropping methods for BLIP2, illustrated in~\cref{fig:methods}, whose goal is to find the \textbf{approximate region of interest} in images, i.e. the region containing the subjects of a question, and then to zoom into that region via visual cropping. One potential drawback of visual cropping is that some questions might need to have a global view of the image. To address this issue, we utilize the fact that MLLMs typically convert the image into a series of tokens. This allows us to directly extend the original image tokens by concatenating the visually cropped image tokens, as illustrated in~\cref{fig:methods}. We use this image token concatenation when applying the visual cropping methods to BLIP-2 models.

\subsection{External Knowledge Visual Cropping}
\label{sec:external}

In this section, we present three automatic question-guided localization methods based on popular off-the-shelf vision-based models, namely CLIP~\cite{clip}, YOLO~\cite{yolo}, and SAM~\cite{kirillov2023segment}. These three methods utilize external vision-based knowledge for the localization process through multimodal encoding, object detection, and semantic segmentation, respectively.

\textbf{\clip.} The intuition of this method is to progressively refine the image towards the region of highest relevance to a given question using CLIP~\cite{clip}. CLIP consists of an image encoder and a text encoder, which are trained on a large dataset of image-caption pairs to map each image (caption) close to its caption (image) and far from all other captions (images). The result is an aligned shared space where various images can be directly compared with various texts. To find the region of interest, given an image-question pair, we first crop the image from the four sides (top, bottom, left, and right) at a cropping ratio of 0.9 to produce four overlapping cropped images. We then use CLIP to assess the semantic similarity between these cropped images and the question. The highest-scoring crop is chosen as the input for the next iteration. This process is repeated for 20 iterations, and the cropped image with the highest CLIP similarity to the question is selected.

\textbf{\yolo.} Instead of a progressive approach to finding the region of interest, in this method we select candidate regions based on a state-of-the-art object detection method: YOLOv8~\citep{Jocher_YOLO_by_Ultralytics_2023} pretrained on COCO~\cite{lin2014mscoco}. Using YOLO, we filter out regions that contain no salient objects -- \ie, regions for which CLIP could mistakenly assign high similarity. More concretely, for each question-image pair, we first use YOLO to collect bounding boxes for all predicted objects with confidence higher than 0.25 (the recommended default).\footnote{https://docs.ultralytics.com/modes/predict} Then, for each predicted bounding box, we crop its corresponding image and compute its similarity to the question using CLIP. Finally, the bounding box with the highest similarity score is selected as the region of interest.

\textbf{\sam.} A limitation of YOLO is that it only provides bounding boxes corresponding to a fixed number of object classes. To overcome this issue, we use the segment anything model (SAM)~\cite{kirillov2023segment}, which has shown state-of-the-art zero-shot segmentation performance. SAM can provide an extensive set of segmentation masks for each image, thus providing a more granular set of salient candidate regions compared to YOLO. More concretely, for each image-question pair, we feed the image into SAM, which provides an extensive set of segmentation masks corresponding to all objects and object parts. Then, we translate these masks into bounding boxes by computing the smallest bounding box that covers each segmentation mask. Finally, the bounding box with the highest CLIP similarity to the question is selected as the region of interest.

\begin{figure*}[!ht]
    \centering
    \includegraphics[trim=0 0 0 0, clip, width=0.99\textwidth]{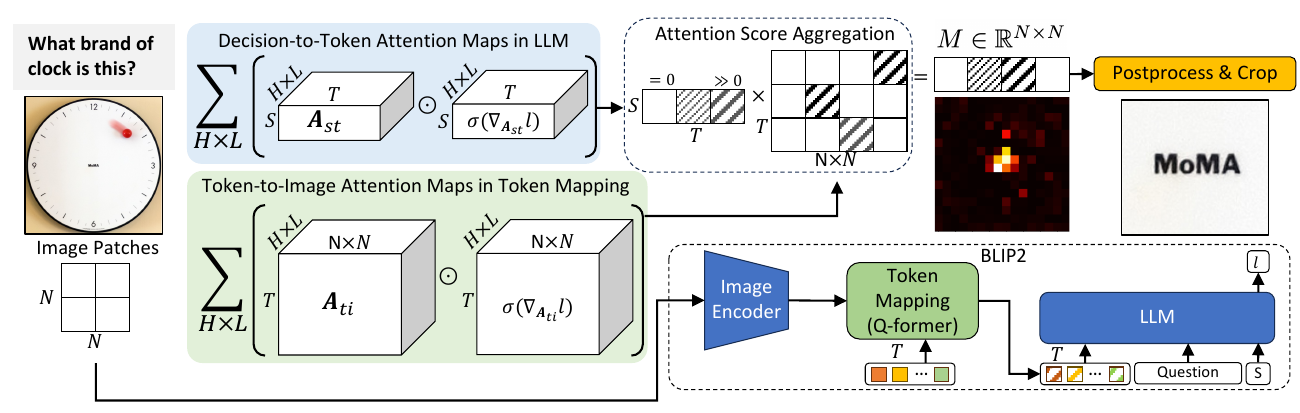}
    \caption{Illustration of \att{} method, where $\sigma$ denotes $ReLU$, $H, L$ the heads and layers of the Transformer, $T$ the query tokens in Q-former, $s$ the last input token for LLM which is used to compute the $l$.}
    \label{fig:cas}
\end{figure*}
\subsection{Native Visual Cropping}
\label{native}

The visual cropping methods we introduced in \cref{sec:external} use off-the-shelf pretrained models which bring in external knowledge, therefore it is unclear whether these methods help the MLLM better understand the question (know where to look in the image) or better perceive small visual details (solving the limitation we observed in~\cref{sec:human_crop}). To address this ambiguity, we next propose two visual cropping methods that natively utilize the MLLM's inference time decision process, \ie, gradients, and attention, for finding the region of interest in images. Performance gain from these methods would show to what extent the MLLM itself knows `where to look' but fails to perceive small details.

\subsubsection{Tracking Gradients (\grad{})}
\label{sec:grad}

\grad{} inspects the gradient of the model's decision with respect to the image (at the pixel level) and determines the image region with the highest gradient magnitude as the region of interest.
To get a differentiable representation of the model's decision, we define it as the logarithm of the maximum softmax probability at the first answer position. This is represented as \( l = \log(\text{softmax}(\mathbf{Z})_{t^*}) \), where \( t^* \) is the token with the highest logit and \( \mathbf{Z} \in \mathbb{R}^{D_v}\) is the output logit of the LLM's head, with $D_v$ denoting the vocabulary size. This representation emphasizes the most relevant token, \( t^* \), while maintaining the contextual information from the entire vocabulary.
Next, we take the gradient of the loss function \( l \) with respect to the input image, \ie, \( \frac{\partial{l}}{\partial{x}} \in \mathbb{R}^{H \times W \times C} \), where \( \mathrm{x} \in \mathbb{R}^{H \times W \times C} \) represents the image with spatial dimensions \( (H, W) \) and \( C \) color channels. Since we are interested in discovering regions where changes have the strongest effect on the decision, we compute the L2-norm of the gradient across the channel dimension \( \lVert \frac{\partial{l}}{\partial{x}} \rVert_2 \in \mathbb{R}^{H \times W} \), and then aggregate over each image patch of the ViT, resulting in a matrix \( M \in \mathbb{R}^{N\times N} \) where each element of \( M \) reflects the importance of a respective image patch in changing the MLLM decision. 

\subsubsection{Tracking Attention Scores (\att{})}
\label{sec:cas}

Instead of looking for regions whose change causes the largest \emph{change in the model's decision}, as we did in~\cref{sec:grad}, in this section, we want to look for the regions whose features have caused the \emph{current model's decision}. To that end, we construct \att{}, which relies on the attention scores computed inside the Transformer blocks to combine how important each image region is to each Q-former token ($A_{ti}$), with how important each Q-former token is to the model's decision ($A_{st}$). \att{} consists of three steps, illustrated in~\cref{fig:cas}, which we will explain in detail below.

\textbf{Generating Token-to-Image Attention Maps.}
As shown in the bottom right block of~\cref{fig:cas}, BLIP-2 uses a frozen ViT to extract image features, which are used for cross-attention in token mapping (Q-former)~\cite{li2023blip}. In the green block of~\cref{fig:cas}, we extract the cross-attention scores through all layers of the Q-former, denoted as \( A_{ti} \in \mathbb{R}^{L \times H \times T \times N^2} \), where $L, H$ are the number of layers and heads-per-layer in the Q-former, respectively, $T$ is the number of query tokens of the Q-former, $N^2$ denotes number of image patches from ViT. Next, we weight \( \mathrm{A_{ti}}\) with the ReLU-gated derivative of $l$ with respect to each of them, formally denoted as \( \sigma(\nabla_{A_{ti}} l) \), which is shown by the right side cuboids. The weighting by gated gradients serves as a way to diminish attention maps that are not actually used in the final model's decision, similar to~\cite{pnpvqa, pnpvqa2, gradcam}. Then we average the attention maps over layers and heads.

\textbf{Generating Decision-to-Token Attention Maps.}
The outputs of the Q-former serve as soft embeddings for the LLM, which conditions the predictions on the soft embeddings and the question. In order to measure the importance of each vision-based soft embedding in answering the question, we extract the cross-attention scores (encoder-decoder architecture) or self-attention scores (decoder-only architecture), resulting in \( \mathrm{A_{st}} \in \mathbb{R}^{L \times H \times S \times T} \), which is shown by~\cref{fig:cas}'s blue block, where $S$ is the number of tokens where we track the loss gradient $l$ (defined in~\cref{sec:grad}), which is 1 according to our definition. We apply the same defined ReLU-gated derivative score \( \sigma(\nabla_{A_{st}} l) \) to  \( \mathrm{A_{st}} \), and average these scores over layers and heads.

\textbf{Attention Score Aggregation.}
The final importance map is obtained by matrix multiplication between the two weighted attention maps (white block in~\cref{fig:cas}), resulting in \( M \in \mathbb{R}^{N \times N}\), indicating how important each image token is to model's decision. In the next section, we will explain how we transform the resulting $M$ into bounding boxes.

\begin{figure*}[t]
    \centering
    \includegraphics[trim=0 0 0 0, clip, width=0.99\textwidth]{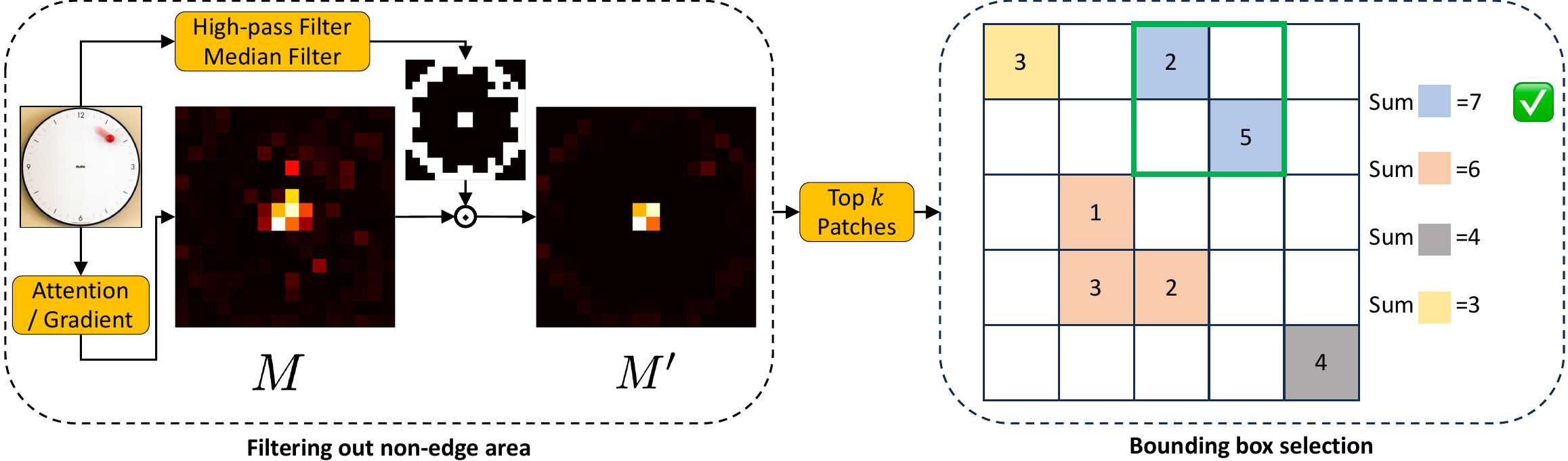}
    \caption{Illustration of the post-processing for the native cropping methods. After retaining the top $k$ image patches, we select the component with the largest sum value. The bounding box is the smallest rectangle containing that component (green box).}
    \label{fig:connect}
\end{figure*}
\subsubsection{Postprocessing and Cropping}
\label{sec:ppc}

Both \att{} and \grad{} result in an importance map \( M \in \mathbb{R}^{N \times N}\). In this section, we will describe how we postprocess $M$ into the final bounding box for visual cropping. We provide extensive visualization and ablation study on the role of our postprocessing in the Appendix.

\textbf{Filtering Out Non-Edge Area.}
Both gradient magnitudes and attention scores sometimes show high values in entirely flat regions (\eg, blue skies). Given that these regions do not contain any visual details, we seek to diminish importance scores on non-edge areas of the image. To that end, we apply a high-pass filter of kernel size \(K_h\) to the image, followed by a median filter of kernel size \(K_m\) to reduce salt-and-pepper noise. The resulting filtered image is then spatially average-pooled into patches to align with $M$'s dimensions, thresholded at its spatial median value to become a binary mask (median is used instead of average to prevent very large/small outlier scores from skewing the threshold), and finally is patch-wise multiplied by $M$ to arrive at an edge-emphasized importance map $M'$.

\textbf{Bounding Box Selection.}
To convert the edge-emphasized importance map $M'$ to a bounding box, we assign a value of 0 to all but the top $k$ patches based on the scores in $M'$, then perform connected component selection (with connectivity considered in all four cardinal directions and the four diagonal directions), and finally, select the connected component with the highest summed score over its patches. The smallest rectangle enclosing this selected connected component is the chosen bounding box for visual cropping.

\section{Experiments}
\label{sec:experiments}

\begin{table*}[t]
\caption{Accuracy of human and automatic visual cropping methods on VQA datasets. For each dataset and model, the best automatic cropping method is depicted in \emph{bold}, and the second-best is \emph{underlined}. \emph{Native} shows whether the bouding box is predicted by tracking MLLM's own inference time dynamics.\\}
\label{tab:main_result}
\centering
\small
\begin{tabular}{lccccccc}
\toprule
Model & Crop Method & Native? & TextVQA & FDVQA & VQAv2 & GQA & AOKVQA\\
\midrule
\multirow{7}{*}{BLIP-2 FlanT5$_\mathrm{XL}$} & w/o cropping & - & 25.91 & 33.94 & 63.43 & 43.85 & 43.42 \\
& \hc & - & 37.68 & 42.29 & - & - & - \\\cmidrule(lr){2-8}
& \clip & \xmark & 30.93 & \bf{36.61} & 63.57 & 45.04 & \bf{45.34} \\
& \yolo & \xmark & 28.94 & 35.87 & 63.39 & \underline{45.20} & 42.72 \\
& \sam & \xmark & \underline{32.31} & \underline{36.33} & \underline{63.85} & \bf{45.23} & 43.00 \\
& \grad & \checkmark & 29.86 & 34.68 & 63.01 & 44.55 & 43.16 \\
& \att & \checkmark & \bf{34.26} & 34.77 & \bf{63.97} & 45.04 & \underline{44.70} \\
\bottomrule
\end{tabular}

\end{table*}

\textbf{Models.} We use two state-of-the-art zero-shot VQA models with open-source code~\cite{li2023blip}: BLIP-2 FlanT5$_\mathrm{XL}$, with an encoder-decoder LLM and, BLIP-2 OPT$_\mathrm{2.3B}$, with a decoder-only LLM. Both models learn to map the input image into soft embeddings that are subsequently processed by a pretrained LLM, with $16 \times 16$ image tokens ($N \times N$) from ViT and $T=32$ query tokens in the Q-former. The results of OPT architecture are reported and discussed in the Appendix, but a notable difference is that visual cropping is less effective for OPT, which we hypothesize is due to complications of concatenation of original and cropped images in this architecture. Implementation details and ablation study of hyperparameters are provided in the Appendix.

\textbf{Datasets.} We consider the validation set of four common VQA datasets and construct a new one tailored towards visual details: 
1) \textbf{FDVQA} is a new dataset that we propose to deliberately focus on small hard-to-perceive visual details. For this purpose, we first selected 400 question-answer pairs of VQAv2~\cite{goyal2017making} on which the zero-shot BLIP-2 model failed to correctly predict the majority of answers in the human annotations, in order to filter out any sample where perception is easy. Then, we collected 3 human annotations per sample identifying whether answering the question requires perceiving small details in the image and whether the model answer is indeed incorrect (\eg, excluding near-synonymous answers or ambiguous questions). Finally, we kept the samples where all 3 annotations agreed, resulting in 109 image-question pairs, and we manually created the ground-truth bounding box around the subject of the question.
2) \textbf{TextVQA}~\cite{textvqa} contains 5,000 questions about textual sequences that appear in 3,166 images, where more than half of the answers require perceiving texts that occupy less than $0.005$ of the total image area. Therefore, TextVQA emphasizes how well a model can read small text, which can serve as a surrogate for how well a model can perceive fine visual details. TextVQA provides Optical Character Recognition (OCR) annotations~\cite{ocr} which we use to approximate the ground-truth answer bounding box for each question by selecting the OCR bounding box containing the text with the highest string similarity with the human-provided answer. This bounding box is used for cropping in \hc. 
3) \textbf{GQA}~\cite{hudson2019gqa} contains 12,578 questions paired with 398 images, using the scene graphs of Visual Genome~\cite{krishna2017visualgenome} to construct highly compositional questions requiring spatial, logical, relational, and comparative reasoning, and explicitly controlling the answer distribution for different groups of questions in order to prevent educated guesses using language and world priors.
4) \textbf{AOKVQA}~\cite{schwenk2022okvqa} contains 1,145 questions about 1,122 images, where the questions require additional knowledge and cannot be answered from the image-question pair alone.
5) \textbf{VQAv2}~\cite{goyal2017making} is a large-scale dataset, a subset of COCO~\cite{lin2014mscoco}, containing 214,354 questions paired with 40,504 images from various objects and settings.

\textbf{Metrics.} We compute zero-shot VQA-score\footnote{https://visualqa.org/evaluation.html} for all benchmarks except GQA. To account for the variability in annotators' answers when measuring the model's accuracy, this accuracy score for any given model answer is defined as $\min(0.3\times n, 1)$, where $n$ denotes the number of times that the model answer appears among the ground-truth answers collected from 10 human annotations. For GQA, we compute accuracy using the official code.\footnote{https://cs.stanford.edu/people/dorarad/gqa/evaluate.html}

\subsection{Results}
\textbf{Visual Cropping Improves Zero-Shot VQA.}
\cref{tab:main_result} shows the accuracy of the proposed visual cropping methods on the five VQA datasets.
First, we consider the detail-focused datasets, FDVQA and TextVQA, where we also have access to human annotations and can report~\hc{} accuracy: we observe that \hc{} improves the accuracy of BLIP-2 FlanT5$_\mathrm{XL}$ by $24.60\%$ on FDVQA and $45.35\%$ on TextVQA, showing the full potential of visual cropping. Among the automatic cropping methods, \clip{} and \att{} reach the closest performance with humans on FDVQA and TextVQA, respectively. 
Next, we consider the more general GQA, AOKVQA, and VQAv2. We observe that visual cropping methods improve the accuracy of the original model BLIP-2 FlanT5$_\mathrm{XL}$. Notably, \sam{} boost the performance of GQA to $45.23\%$, exceeding the largest version of BLIP-2~\cite{li2023blip}, while \att{} reaches a consistent strong performance across the three benchmarks. Thus, the visual cropping accuracy gains on fine details (observed in FDVQA and TextVQA) do not seem to come at the cost of their accuracy on larger visual details and relations.

\begin{figure*}[t]
    \centering
    \includegraphics[trim=0 10 0 0, clip, width=0.99\textwidth]{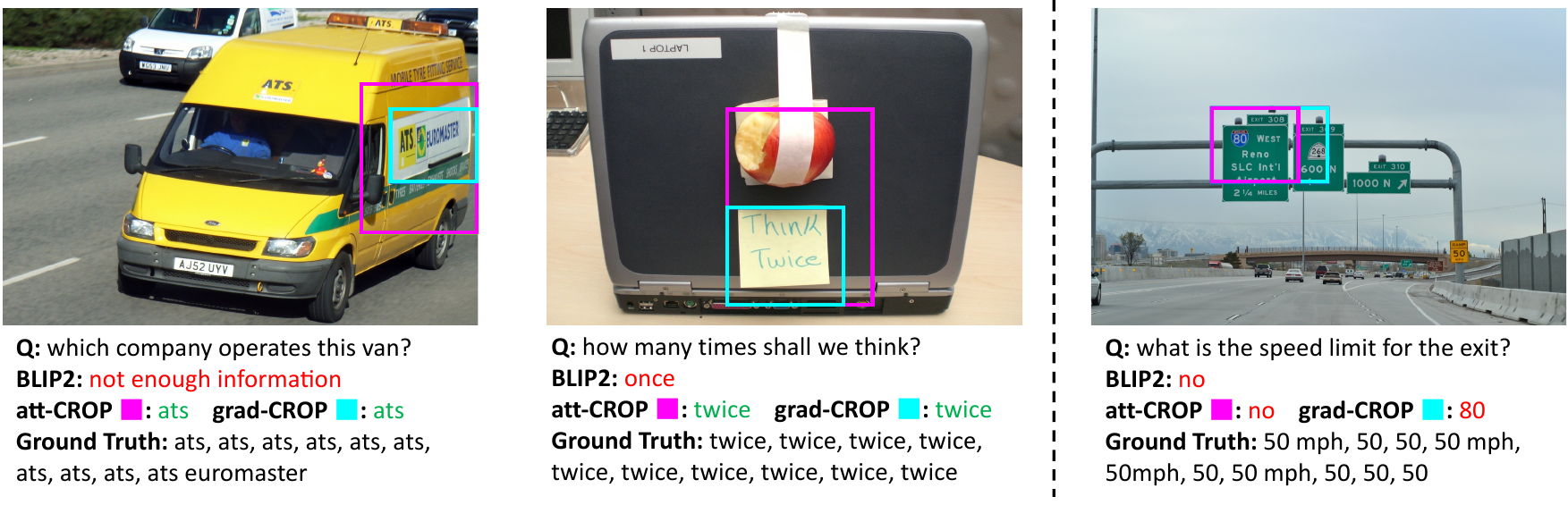}
    \caption{Examples of success and failure of native cropping methods in correcting the mistakes of BLIP-2 FlanT5$_\mathrm{XL}$ on TextVQA. (Left, middle) BLIP-2 is able to recognize the salient area although fail to answer the question, and therefore native cropping is effective. (Right) BLIP-2 entirely fails to recognize the salient area and therefore native cropping is ineffective.}
    \label{fig:crop_examples}
\end{figure*}
\begin{figure}[!ht]
    \centering
    \includegraphics[trim=0 0 0 0, clip, width=0.49\textwidth]{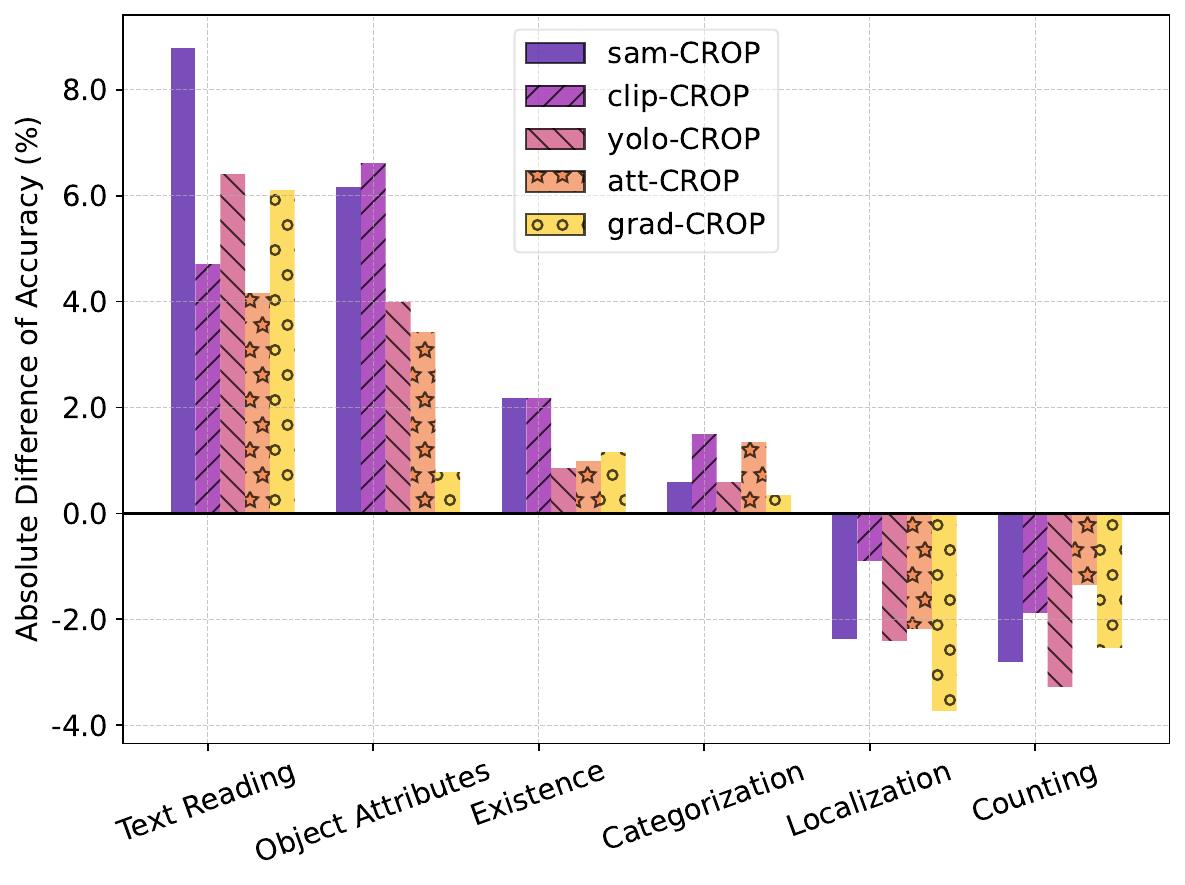}
    \caption{Accuracy gain of visual cropping methods compared to no cropping, when applied to BLIP-2 FlanT5$_\mathrm{XL}$ on different question types in VQAv2. The x-axis is sorted based on the combined gain of all methods. The detailed question types and figures for BLIP-2 OPT are provided in the Appendix.}
    \label{fig:question_type}
    \vskip -1em
\end{figure}

\textbf{Effect of Visual Cropping On Different Question Types.} To gain deeper insights into the granular benefits of visual cropping, \cref{fig:question_type} shows how the proposed visual cropping methods impact the accuracy of zero-shot VQA models on various question types in VQAv2 (the selection of these types is detailed in Appendix). Questions that often concern small visual details, \ie, text reading and object attributes, gain the most from visual cropping, consistent with our findings in FDVQA and TextVQA. However, we observe that questions that require a global view of the image, \ie, localization and counting, become harder to answer as a result of visual cropping. These findings suggest that our mechanism for combining the original and cropped image tokens is not always successful in maintaining the global image information, motivating the development of more complex combination mechanisms (we demonstrate such an attempt in the Appendix, where we use a question-gate that dynamically decides whether to conduct visual cropping or not). Additionally, we observe that native cropping methods, particularly \att{}, demonstrate closer performance to no cropping across different question types. This indicates an alignment with the MLLM’s own judgment in selecting bounding boxes, a result that is intuitive given that the cropping is based on the MLLM's own attention mechanisms.

\textbf{MLLMs Can Know `Where to Look', Even If They Answer Incorrectly.}
As shown in~\cref{fig:crop_examples} (left and middle), despite initially providing incorrect answers, BLIP-2 often focuses on the relevant image areas, indicated by the predicted bounding boxes by native visual cropping methods (the company name and sticky note). In addition, a similar conclusion can be drawn quantitatively from~\autoref{tab:main_result}, where BLIP-2 FlanT5$_\mathrm{XL}$'s performance on TextVQA improved by 45.45\% by adopting \hc{}, and tracking its native attention can approximate 70.94\% of such improvement, outperforming the cropping methods that use external localization methods. The above outcome suggests that even if MLLMs do not answer correctly, they often still identify the salient areas in the images.

\begin{table}[!h]
\caption{Average cropping inference time (in seconds) of the proposed visual cropping methods.\\}
\label{tab:inference_time}
\centering
\begin{tabular}{lcc}
\toprule
 Methods & CPU & GPU \\
\midrule
 \clip  &  5.461 & 1.072 \\
 \yolo & \textbf{0.970} &  0.355 \\
\sam & 91.532 & 3.329 \\
 \grad ~(BLIP-2 FlanT5$_\mathrm{XL}$) & 9.486  & 0.851 \\
  \grad ~ (BLIP-2 OPT$_\mathrm{2.3B}$) & 9.488 & 0.824 \\
 \att~ (BLIP-2 FlanT5$_\mathrm{XL}$) & 3.464  & 0.447 \\
 \att~ (BLIP-2 OPT$_\mathrm{2.3B}$)  & 3.610 & \textbf{0.298} \\
\bottomrule
\end{tabular}
\end{table}

\textbf{Time Overhead of Visual Cropping.} In~\cref{tab:inference_time}, we report the average inference time of the proposed visual cropping methods on GPU (NVIDIA RTX A5000) and CPU (Intel Xeon Gold 5215 2.50GHz).
When conducting inference on a GPU, both the \att{} and \grad{} methods exhibit speeds comparable to \yolo{}, which is optimized for time efficiency. This result implies that tracking the inference time attention and gradient of multimodal LLMs is not only effective in accurately identifying the salient areas of an image but also maintains a high processing speed. When only CPU is available, \yolo{} provides the best accuracy and time trade-off.

\section{Conclusion and Future Works}
\label{sec:discussion}

In this work, we qualitatively and quantitatively showed the limitation of a state-of-the-art zero-shot VQA MLLM, namely BLIP-2, in perceiving small visual details, and then proposed five automatic visual cropping methods as potential inference time solutions to mitigate this limitation. Our findings suggest that MLLMs should be used with caution in detail-sensitive VQA applications, and that visual cropping is a promising direction to improve their zero-shot performance. Finally, our findings reveal several directions for future research: 1) while we showed the limitation in perceiving details exists, its cause remains unknown; 2) as we expect the limitation to affect other MLLMs beyond BLIP-2, extending this study to other models released in the future is valuable; and, 3) while our proposed automatic visual cropping methods improve the accuracy on small visual details, they are still not as successful as human visual cropping, which encourages new visual cropping methods.

\clearpage
\section*{Impact Statement}
This paper presents work whose goal is to advance the field of Machine Learning. There are many potential societal consequences of our work, none of which we feel must be specifically highlighted here.

\bibliography{ref}
\bibliographystyle{icml2024}

\newpage
\appendix
\onecolumn
\section{Implementation Details.}
We use \textit{python 3.8.16, salesforce-lavis 1.0.2, transformers 4.29.1 and torch 2.0.1} for all the experiments. Our environment consists of an Intel(R) Xeon(R) Gold 5215 CPU @ 2.50GHz with 40 cores and 256 GB of RAM, and we utilize NVIDIA RTX A5000 GPUs for our experiments. For hyperparameters of~\cref{sec:ppc}, we use \(K_h\)=3, \(K_m\)=7, and \(k\)=30.

\section{Result for BLIP-2 OPT model}
The result of BLIP2-OPT$_\mathrm{2.3B}$ is shown in~\autoref{tab:opt}.
We observe that visual cropping methods are not as effective as for FlanT5, and can even cause a slight decline in the overall accuracy in VQAv2 and GQA.
We hypothesize that the observations are linked to the pretraining objectives of two different LLMs. Unlike FlanT5, which benefits from instruction finetuning, the OPT model was not trained with such an objective. As is shown by~\cite{flant5}, instruction finetuning is crucial for understanding inputs that are different from what the model was originally trained on, such as handling different lengths of visual embeddings.
\begin{table*}[!h]
\caption{Accuracy of human and automatic visual cropping methods on VQA datasets. For each dataset and model, the best automatic cropping method is depicted in \emph{bold}, and the second-best is \emph{underlined}. \emph{External} shows whether any external model beyond the under-study zero-shot BLIP-2 is used during inference.\\}
\label{tab:opt}
\centering
\small

\begin{tabular}{lccccccc}
\toprule
Model & Crop Method & Native? & TextVQA & FDVQA & VQAv2 & GQA & AOKVQA\\
\midrule
\multirow{7}{*}{BLIP-2 OPT$_\mathrm{2.3B}$} & w/o cropping & - & 23.93 & 35.14 & 51.22 & 31.95 & 31.57 \\
& \hc & - & 31.21 & 42.11 & - & - & - \\\cmidrule(lr){2-8}
& \clip & \xmark & 26.35 & \bf{35.60} & \bf{49.67} & \bf{31.14} & \bf{32.69} \\
& \yolo & \xmark & 25.27 & 34.86 & \underline{48.63} & 30.39 & 29.43 \\
& \sam & \xmark & \underline{26.45} & \bf{35.60} & 48.58 & 30.47 & \underline{31.07} \\
& \grad & \checkmark & 25.07 & 35.23 & 47.61 & 30.28 & 28.99 \\
& \att & \checkmark & \bf{26.77} & 34.04 & 47.68 & \underline{31.08} & 30.28 \\
\bottomrule
\end{tabular}

\end{table*}

\clearpage

\section{Selective Prediction Based on Question Types}

\begin{table*}[!h]
    \centering
    \caption{We select 6 question types from~\textbf{VQAv2} based on their first two words to study the granular accuracy of visual cropping methods in~\autoref{sec:experiments}. The total number of instances per question type is reported in the last row, with 140691 instances belonging to none.\\}
    \label{tab:question_types}
    \begin{tabular}{cccccc}
    \toprule
    \multicolumn{6}{c}{\textbf{Question types}} \\
    \textbf{Reading} & \textbf{Object Attributes} & \textbf{Existence} & \textbf{Categorization} & \textbf{Localization} & \textbf{Counting} \\
    \midrule
    what letter & what pattern & is anyone & what street & where is & how many \\
    what brand & what color & is there & what direction & where are & how much \\
    & what breed & are there & what animal & where was & \\
    & what colors & is that & what fruit & & \\
    & what style & are all & what vegetable & & \\
    & what material & is everyone & what food & & \\
    & what shape & is one & what game & & \\
    & & is she & what sport & & \\
    & & is he & & & \\
    \midrule
    1064 & 22053 & 16426 & 4168 & 6329 & 23623 \\
    \bottomrule
    \end{tabular}
\end{table*}
	
\definecolor{Gray}{gray}{0.9}
\begin{table*}[!h]
\caption{Accuracy after we apply the question type selection gate to all the datasets, the accuracy after the gate is shown below, before the gate is shown above.\\}
\label{tab:selective_result}
\centering
\small

\begin{tabular}{lccccccc}
\toprule
Model                             & Crop Method            & Native?                     & TextVQA & FDVQA  & VQAv2 & GQA   & AOKVQA \\
\midrule
\multirow{14}{*}{BLIP-2 FlanT5$_\mathrm{XL}$} 
                                  & w/o cropping           & -                           & 25.91   & 33.94  & 63.43 & 43.85 & 43.42  \\
                                  & \hc                    & -                           & 37.68   & 42.29  & -     & -     & -      \\\cmidrule(lr){2-8}
                                  & \multirow{2}{*}{\clip} & \multirow{2}{*}{\xmark}     & 30.93   & 36.61  & 63.63 & 45.04 & 45.34  \\
                                  &                        &                             & 30.53   & 35.69  & 64.10 & 45.06 & 45.55  \\\cmidrule(lr){4-8}
                                  & \multirow{2}{*}{\yolo} & \multirow{2}{*}{\xmark}     & 28.94   & 35.87  & 63.39 & 45.20 & 42.72  \\
                                  &                        &                             & 28.77   & 36.24  & 63.82 & 45.21 & 43.21  \\\cmidrule(lr){4-8}
                                  & \multirow{2}{*}{\sam}  & \multirow{2}{*}{\xmark}     & 32.31   & 36.33  & 63.85 & 45.23 & 43.00  \\
                                  &                        &                             & 32.11   & 38.53  & 64.23 & 45.24 & 43.17  \\\cmidrule(lr){4-8}
                                  & \multirow{2}{*}{\grad} & \multirow{2}{*}{\checkmark} & 29.86   & 34.68  & 63.01 & 44.55 & 43.16  \\
                                  &                        &                             & 33.99   & 34.31  & 64.25 & 45.05 & 45.02  \\\cmidrule(lr){4-8}
                                  & \multirow{2}{*}{\att}  & \multirow{2}{*}{\checkmark} & 34.26   & 34.77  & 63.97 & 45.04 & 44.70  \\
                                  &                        &                             & 33.83   & 35.23  & 64.30 & 45.21 & 44.76  \\
\toprule
\multirow{14}{*}{BLIP-2 OPT$_\mathrm{2.3B}$} 
                                  & w/o cropping           & -                           & 23.93   & 35.14  & 51.22 & 31.95 & 31.57  \\
                                  & \hc                    & -                           & 31.21   & 42.11  & -     & -     & -      \\\cmidrule(lr){2-8}
                                  & \multirow{2}{*}{\clip} & \multirow{2}{*}{\xmark}     & 26.35   & 35.60  & 49.67 & 31.14 & 32.69  \\
                                  &                        &                             & 26.21   & 36.79  & 49.99 & 31.14 & 32.75  \\\cmidrule(lr){4-8}
                                  & \multirow{2}{*}{\yolo} & \multirow{2}{*}{\xmark}     & 25.27   & 34.86  & 48.63 & 30.39 & 29.43  \\
                                  &                        &                             & 25.32   & 36.06  & 49.27 & 30.40 & 29.93  \\\cmidrule(lr){4-8}
                                  & \multirow{2}{*}{\sam}  & \multirow{2}{*}{\xmark}     & 26.45   & 35.60  & 48.58 & 30.47 & 31.07  \\
                                  &                        &                             & 26.27   & 37.43  & 49.18 & 30.49 & 31.19  \\\cmidrule(lr){4-8}
                                  & \multirow{2}{*}{\grad} & \multirow{2}{*}{\checkmark} & 25.07   & 35.23  & 47.61 & 30.28 & 28.99  \\
                                  &                        &                             & 24.99   & 36.42  & 48.18 & 30.31 & 29.20  \\\cmidrule(lr){4-8}
                                  & \multirow{2}{*}{\att}  & \multirow{2}{*}{\checkmark} & 26.77   & 34.04  & 47.68 & 31.08 & 30.28  \\
                                  &                        &                             & 26.70   & 35.14  & 48.30 & 31.09 & 30.48  \\
\bottomrule
\end{tabular}

\end{table*}

As is depicted by \autoref{tab:question_types}, we categorize the questions based on their initial two words. Inspired by the performance variance shown by~\autoref{fig:question_type}, we design a gate that dynamically determines whether to conduct visual cropping, based on the question type. Specifically, we opt for answers without cropping in the \textit{Counting} and \textit{Localization} group since cropping diminishes the performance, and select cropping answers for the rest. It is important to note that although the observation is from VQAv2, this strategy is not specific to a single dataset but is rather applied universally across all five datasets examined in our study. As is shown in~\ref{tab:selective_result}, ~\textbf{VQAv2} and ~\textbf{AOKVQA} benefit from the strategy the most.

\clearpage

\section{Sensitivity to Hyperparameter Values}

\begin{figure*}[!h]
    \centering
    \includegraphics[trim=0 0 0 0, clip, width=0.85\textwidth]{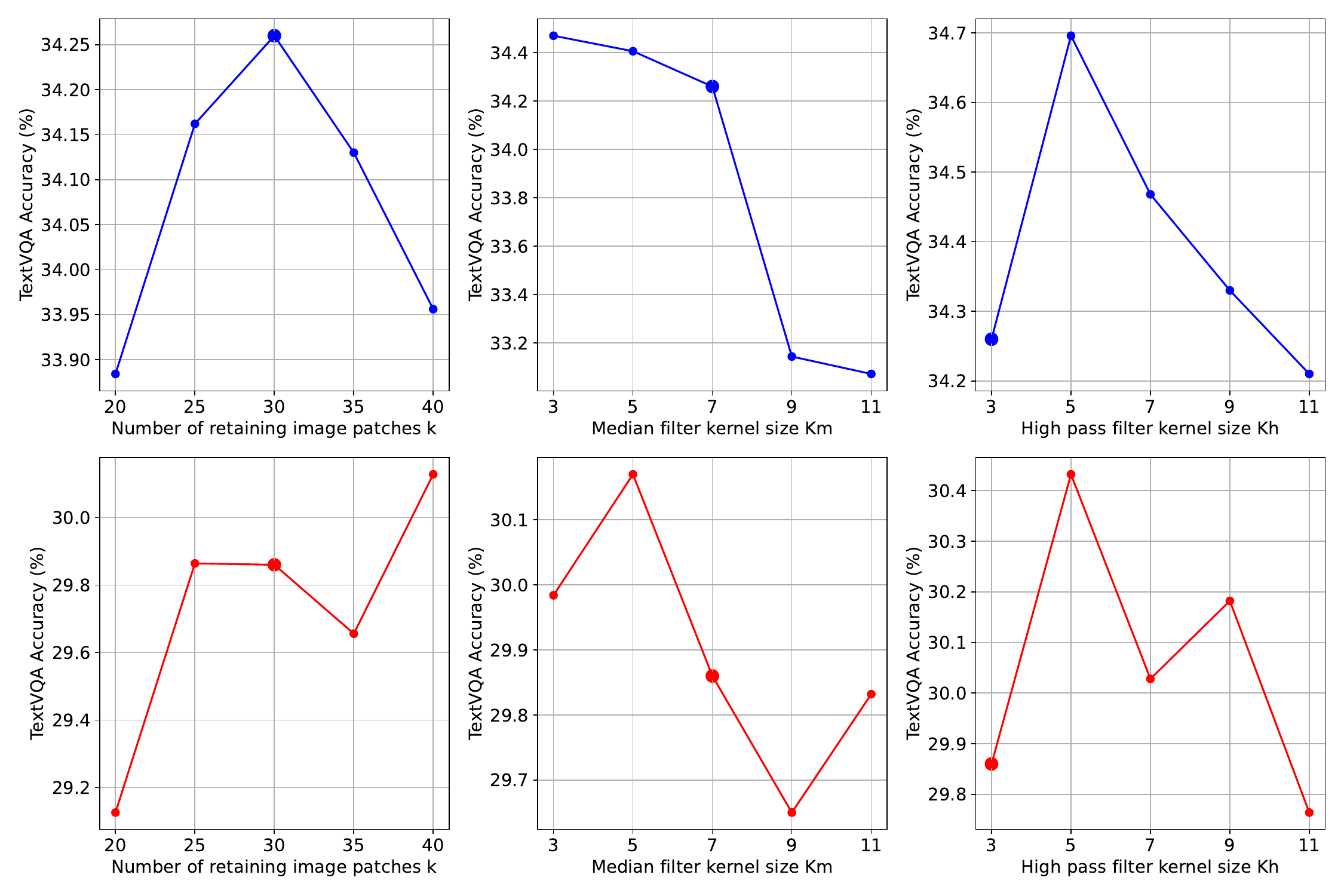}
    \caption{The sensitivity of the hyperparameters \(k\), \(K_h\) and \(K_m\) introduced in~\cref{sec:ppc} of~\att{} (top) and~\grad{} (bottom) on BLIP-2 FlanT5$_\mathrm{XL}$, conducted on~\textbf{TextVQA} dataset. In studying each hyperparameter, the other hyperparameters are kept at their default value that we use in the main paper (illustrated with enlarged markers).}
    \label{fig:ablation}
\end{figure*}

\begin{table}[!h]
\caption{Accuracy comparison of \att{} and \grad{} with and without high-pass filter on~\textbf{TextVQA} dataset.\\\\}
\label{tab:ablation}
\centering
\small
\begin{tabular}{lcc}
\toprule
 Methods & w/ High-Pass Filter & w/o High-Pass Filter \\
\midrule
 \grad  &  29.86 & 26.20 \\
 \att & 34.26 & 32.36 \\
\bottomrule
\end{tabular}
\end{table}

We study the hyperparameter sensitivity of \(K_h\), \(K_m\), and \(k\) introduced in~\autoref{sec:ppc}. We observe in~\autoref{fig:ablation}, that the both \att{} and \grad{} methods are robust to the choice of hyperparameters (less than $2\%$ absolute change in accuracy within a reasonable range), with the number of retaining patches ($k$) having the least sensitivity, and median filter kernel size ($K_h$) having the largest sensitivity. This suggests that both methods' localization is susceptible to salt and pepper noise (\ie, spuriously high values at individual locations) which we target with the median filter.

Furthermore, to study the importance of the high-pass filter as a whole, we remove the entire high-pass filter and test on~\textbf{TextVQA}. We observe in~\autoref{tab:ablation} that the high-pass filter is indeed important for both methods, and that it affects the performance of \grad{} the most. This suggests that \grad{} is more susceptible to assigning spurious high values to constant (non-edge) regions in the image, which is consistent with our expectation: \grad{} uses pixel-level gradients for localization which can have spurious large values because infinitesimal changes in constant image regions can create edges that result in a spike in the model's decision; in contrast, \att{} uses the model's attention maps to localize regions that have resulted in the current model's decision, making it less susceptible to how infinitesimal changes in the input can change the decision.

\clearpage

\section{Additional Examples on Model's Predictions}

\begin{figure*}[t]
    \centering
    \includegraphics[trim=0 0 0 0, clip, width=0.95\textwidth]{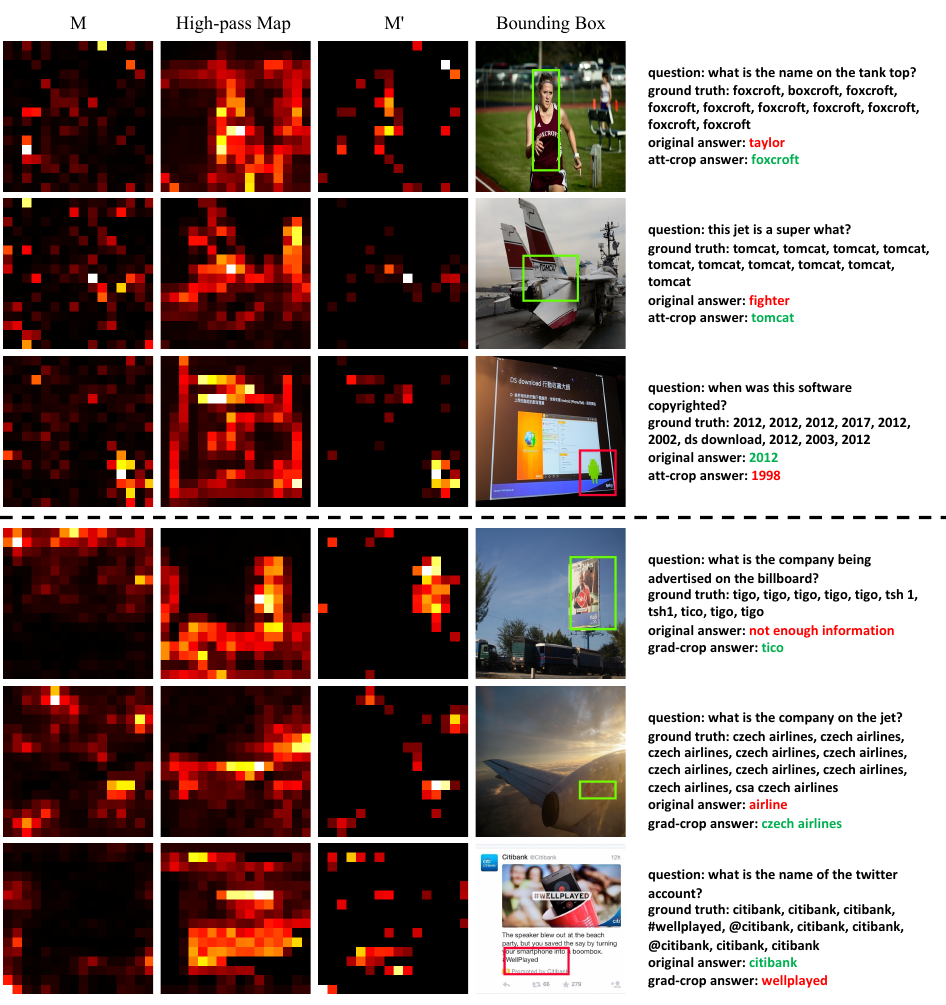}
    \caption{\textbf{TextVQA} success (green bounding boxes) and failure (red bounding boxes) examples of \att{} (top) and \grad{} (bottom). We also visualize the $M$ (importance map before high-pass), high-pass filter map, and $M'$ (importance map after high-pass) as defined in \ref{sec:ppc} to illustrate the details.}
    \label{fig:exampletextvqa}
\end{figure*}

\begin{figure*}[t]
    \centering
    \includegraphics[trim=0 0 0 0, clip, width=0.95\textwidth]{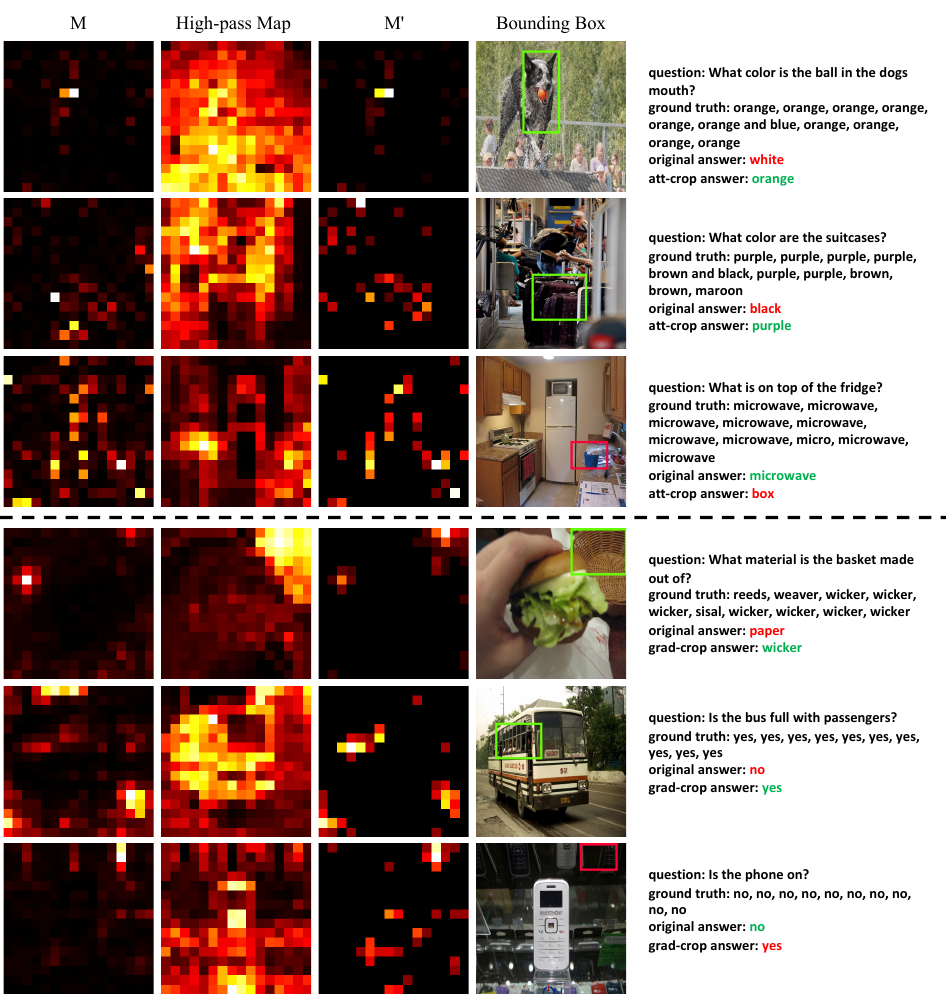}
    \caption{\textbf{VQAv2} success (green bounding boxes) and failure (red bounding boxes) examples of \att{} (top) and \grad{} (bottom). We also visualize the $M$ (importance map before high-pass), high-pass filter map, and $M'$ (importance map after high-pass) as defined in \ref{sec:ppc} to illustrate the details.}
    \label{fig:examplevqav2}
\end{figure*}

We show additional examples of our native cropping~\att{} and~\grad{} on~\textbf{TextVQA}, ~\textbf{GQA},~\textbf{AOKVQA} and~\textbf{VQAv2}, as well as the visualization of their importance maps before high-pass filtering ($M$) and after ($M'$).
We would like to highlight the following qualitative observations: 

\begin{enumerate}
    \item High-pass filter is helping~\grad{} more than~\att{}, since the difference between $M$ and $M'$ is larger in~\grad{} than in~\att{}. This is consistent with the qualitative analysis in~\autoref{tab:ablation}.

    \item In addition to the largest connected component, both~\att {} and~\grad{} often focus on multiple regions, including some irrelevant regions. For example, two regions on the sky in the fifth example of~\autoref{fig:exampletextvqa}, and the focus on the finger in the fourth example of~\autoref{fig:examplevqav2}.

    \item We observe that incorrect cropping can mislead the MLLM even if the original prediction is correct.
\end{enumerate}

\begin{figure*}[t]
  \centering

  \begin{subfigure}[t]{0.32\textwidth}
    \centering
    \includegraphics[width=\linewidth, height=3cm]{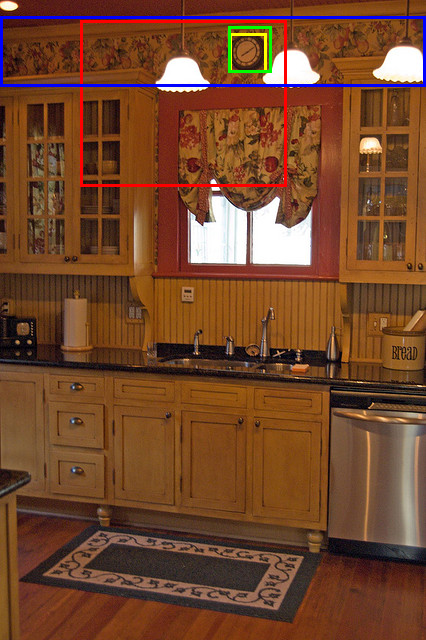}
    \caption{Question: What time is it?}
    \label{SF1}
  \end{subfigure}
  \hfill
  \begin{subfigure}[t]{0.32\textwidth}
    \centering
    \includegraphics[width=\linewidth, height=3cm]{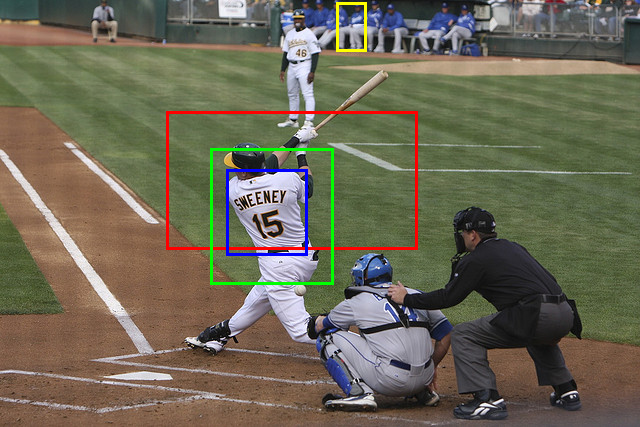}
    \caption{Question: What is the name on back of the players Jersey?}
    \label{SF2}
  \end{subfigure}
  \hfill  
  \begin{subfigure}[t]{0.32\textwidth}
    \centering
    \includegraphics[width=\linewidth, height=3cm]{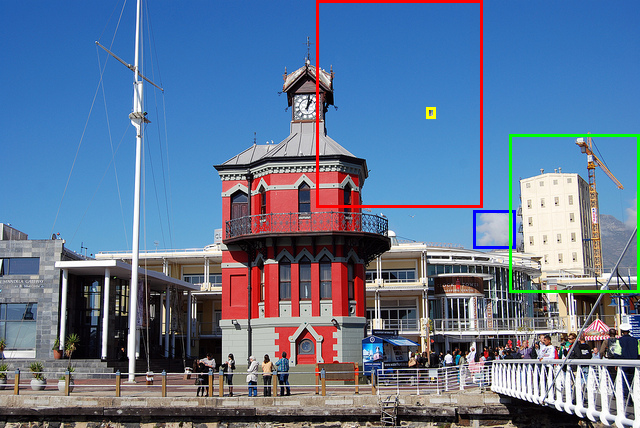}
    \caption{ Question: Is there a mountain in background behind white building?}
    \label{SF3}
  \end{subfigure}

  \caption{Success and failure examples of the \textbf{external} visual cropping methods: \color{black} \clip~ (\color{red} $\bullet$\color{black}), \yolo~(\color{yellow} $\bullet$\color{black}), \sam~\color{black} (\color{blue} $\bullet$\color{black}),
  and 
  \color{black} \hc~(\color{green} $\bullet$\color{black}).}
  \label{fig:external_examples}
\end{figure*}

We also provide three examples of external cropping in~\autoref{fig:external_examples}.

\begin{figure*}[t]
    \centering
    \includegraphics[trim=0 0 0 0, clip, width=0.95\textwidth]{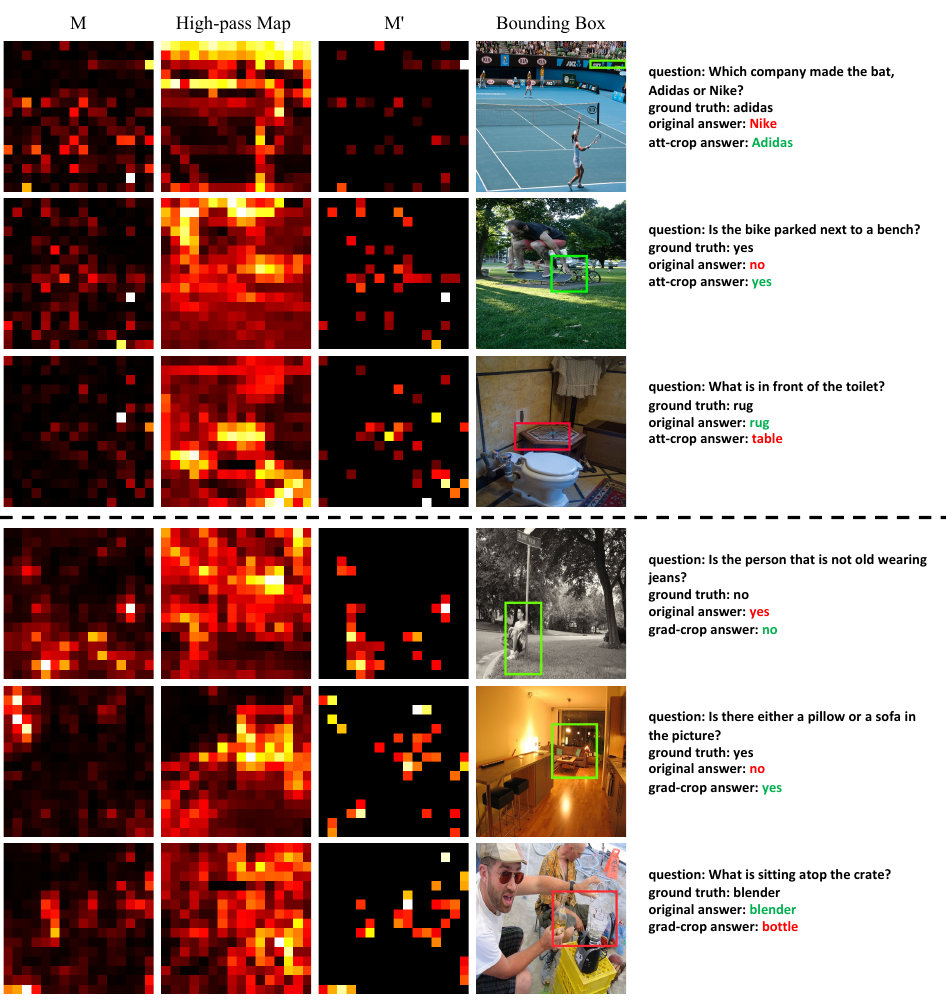}
    \caption{\textbf{GQA} success (green bounding boxes) and failure (red bounding boxes) examples of \att{} (top) and \grad{} (bottom). We also visualize the $M$ (importance map before high-pass), high-pass filter map, and $M'$ (importance map after high-pass) as defined in \ref{sec:ppc} to illustrate the details.}
    \label{fig:examplegqa}
\end{figure*}

\begin{figure*}[t]
    \centering
    \includegraphics[trim=0 0 0 0, clip, width=0.95\textwidth]{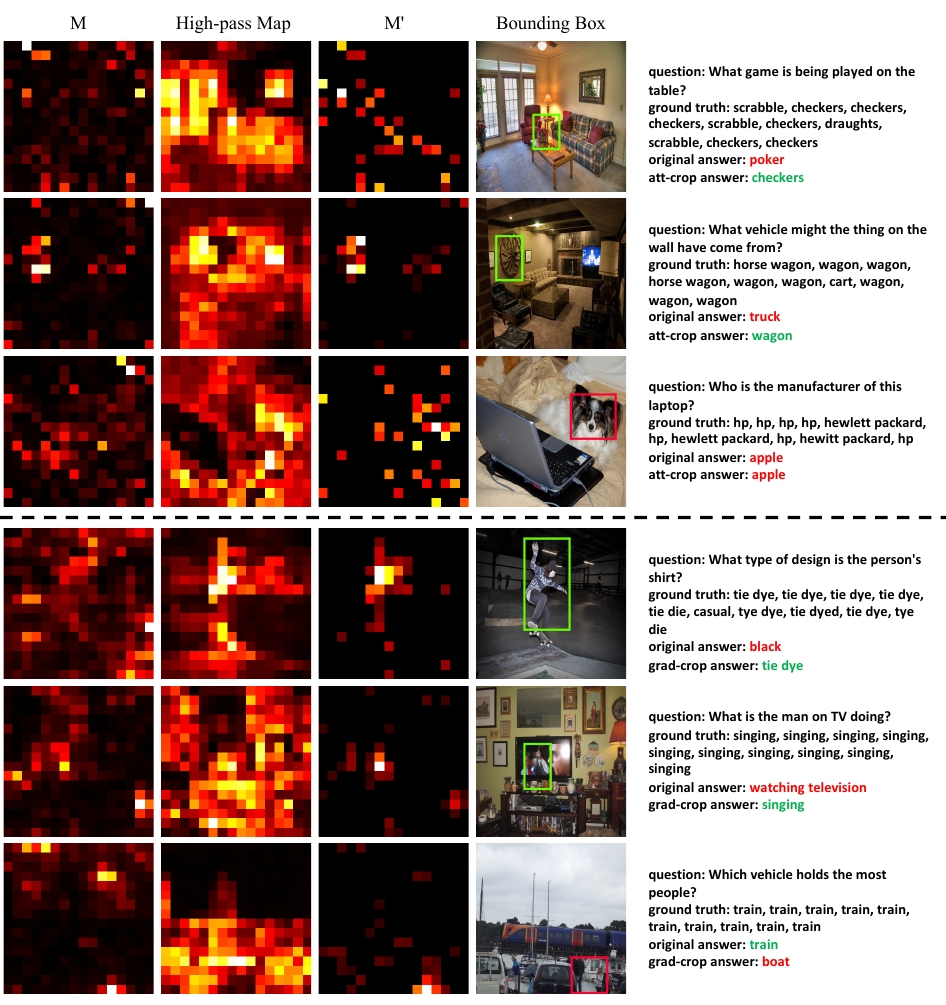}
    \caption{\textbf{AOKVQA} success (green bounding boxes) and failure (red bounding boxes) examples of \att{} (top) and \grad{} (bottom). We also visualize the $M$ (importance map before high-pass), high-pass filter map, and $M'$ (importance map after high-pass) as defined in \ref{sec:ppc} to illustrate the details.}
    \label{fig:exampleaokvqa}
\end{figure*}

\end{document}